\documentclass[conference]{IEEEtran}

\IEEEoverridecommandlockouts
\usepackage{graphicx} 
\usepackage{pifont}
\usepackage{amsmath}
\usepackage{makecell}
\usepackage{amssymb}  
\usepackage{amsfonts}
\usepackage{algorithmic}
\usepackage{array}
\usepackage{multicol}
\usepackage{txfonts}
\usepackage{textcomp}
\usepackage{stfloats}
\usepackage{url}
\usepackage{verbatim}
\usepackage{graphicx}
\usepackage{subfigure}
\usepackage{caption}
\usepackage{subcaption}
\usepackage[ruled, vlined, linesnumbered]{algorithm2e}
\usepackage{cite}
\usepackage{booktabs}
\usepackage{amssymb,amsfonts,bm}
\usepackage{amsmath,tikz}
\usepackage{color}
\usepackage{mathtools}
\usepackage[hidelinks]{hyperref} 
\usepackage{rotating}
\usepackage{blkarray}
\usepackage{physics}
\usepackage{etoolbox}
\usepackage{cuted}
\usepackage{etoolbox}
\AfterEndEnvironment{strip}{\leavevmode}

\usetikzlibrary{arrows}
\usepackage{makecell}

\usepackage{comment}
\usepackage{multirow}
\usepackage{geometry}
\usepackage{caption}

\begin{document}
%\title{Handle Object Navigation as Weighted Traveling Repairman Problem}
\title{Global Planning for Object Navigation via a Weighted Traveling Repairman Problem Formulation}

\author{Ruimeng Liu,
        Xinhang Xu,
        Shenghai Yuan,~\IEEEmembership{Member,~IEEE}
        And Lihua Xie,~\IEEEmembership{Fellow,~IEEE}% <-this % stops a space

        \thanks{This work is supported by the National Research Foundation of Singapore under its Medium-Sized Center for Advanced Robotics Technology Innovation.}
\thanks{All authors are with the Centre for Advanced Robotics Technology Innovation (CARTIN), School of Electrical and Electronic Engineering, Nanyang Technological University, 50 Nanyang Avenue, Singapore 639798, { \{shyuan,elhxie\}@ntu.edu.sg}.}
}

% Ruimeng Liu 449925
% xinhang xu 353500
% yuan sheghai 185927
% lihua xie 115410

% % The paper headers
% \markboth{Journal of \LaTeX\ Class Files,~Vol.~14, No.~8, August~2015}%
% {Shell \MakeLowercase{\textit{et al.}}: Bare Demo of IEEEtran.cls for IEEE Journals}

\maketitle

\begin{strip}
    \centering
    \vspace{-20mm}
    \includegraphics[width=\linewidth]{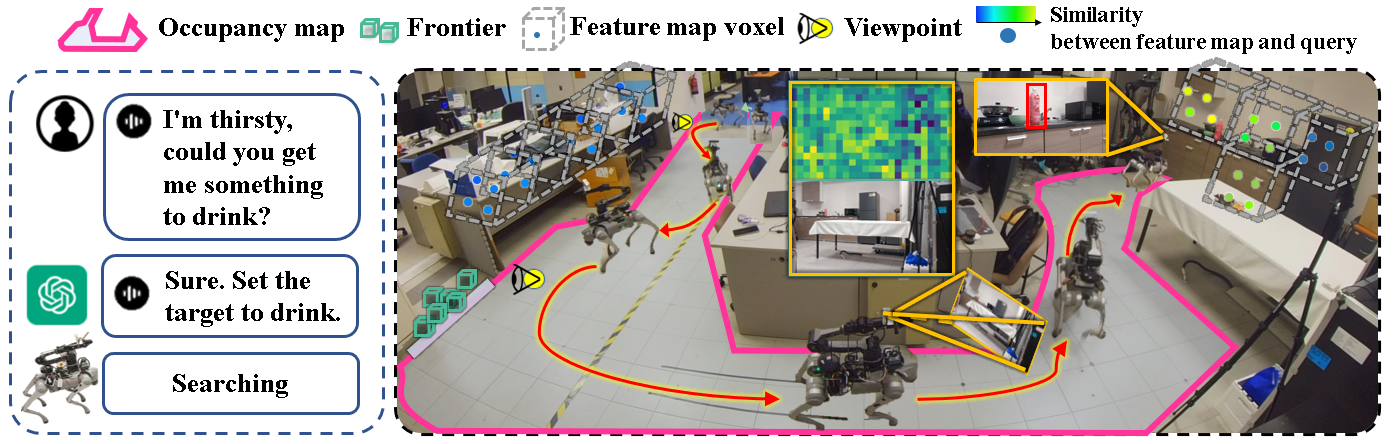}
    \vspace{-7mm}
    \captionof{figure}{ Implementation of our approach in a real robot. A VLM would extract features from rgb image and compute similarity with the target. Features from the VLM are stored and fused in 3D voxels and would match with the embeddings of the query. The right of the figure is a pantry area, so it presents higher similarity than office area in the left. Frontiers are generated in the margin of free space and unknown space. Viewpoints from frontiers guide the robot to enlarge the explored area.
    }
    \label{fig:teaser}
    \vspace{-3mm}
\end{strip}

\IEEEpeerreviewmaketitle

% \begin{figure*}[H]
%     \centering
%     \includegraphics[width=0.90\linewidth]{fig/real_diagram.png}
%     \caption{Implementation of our approach in a real robot. The left image visualizes items that support the search framework. The right top image represents the detection result of a camera in the robot. The right bottom image is a third view.}
%     \label{fig:real}
% \end{figure*}

\begin{abstract}
Zero-Shot Object Navigation (ZSON) requires agents to navigate to objects specified via open-ended natural language without predefined categories or prior environmental knowledge. While recent methods leverage foundation models or multi-modal maps, they often rely on 2D representations and greedy strategies or require additional training or modules with high computation load, limiting performance in complex environments and real applications. We propose WTRP-Searcher, a novel framework that formulates ZSON as a Weighted Traveling Repairman Problem (WTRP), minimizing the weighted waiting time of viewpoints. Using a Vision-Language Model (VLM), we score viewpoints based on object-description similarity, projected onto a 2D map with depth information. An open-vocabulary detector identifies targets, dynamically updating goals, while a 3D embedding feature map enhances spatial awareness and environmental recall. WTRP-Searcher outperforms existing methods, offering efficient global planning and improved performance in complex ZSON tasks. Code and design will be open-sourced upon acceptance. 
%Code and more demos will be avaliable on \url{https://github.com/lrm20011/WTRP_Searcher}.

\end{abstract}

% \begin{IEEEkeywords}
% Scene representation, Mapping, Path planning.
% \end{IEEEkeywords}

\section{Introduction}

Zero-Shot Object Navigation (ZSON) is a challenging task in robotics and AI, where an agent is required to navigate to specified targets using open-ended natural language. Unlike conventional object navigation methods \cite{kiss2025sampling,kostyukov2025global,bouhabza2025energy,shang2025hgee,pan2025action,liu2025enhanced,analooee2025dmtm,yang2025fast}, ZSON operates without a predefined set of object categories or prior knowledge of the environment, making it more adaptable to real-world applications. 

Object navigation has received considerable research attention in recent years. Earlier works \cite{wijmans2019dd,chang2020semantic} implicitly utilize the comprehension of the semantic environment through reinforcement or imitation learning but are limited to closed-set settings and certain environments. Recent training-free methods \cite{yu2023l3mvn,zhou2023esc,dorbala2023can} leverage large language models (LLM) for planning by converting observations into language queries. However, this approach is inefficient and neglects complex 3D correspondence, leading to sub-optimal decisions. Many experiments in biology \cite{golledge1999wayfinding} and robotics \cite{chaplot2020neural,chaplot2020learning} have shown that spatial representation and memory are significant mechanisms in human navigation, and explicit maps improve performance and efficiency over end-to-end methods in various navigation tasks. More and more researchers are using prior map-based methods \cite{chaplot2020object}. They build semantic maps \cite{ huang2023visual, wei2024ovexp, lukas2024one,yu2024vln} for further inferring or decision making by close-set detector or open-vocabulary Vision-Language Model (VLM). Some works also combined this with a traditional frontier method, which marks the boundary of observed and unobserved areas for more efficient navigation \cite{yu2023frontier,yokoyama2024vlfm}.

Despite these advancements, several challenges remain in map-based and frontier-based approaches. Many methods \cite{huang2023visual, lukas2024one} employ top-down 2D feature maps, which struggle with overlapping objects and occluded environments (e.g., objects inside cabinets). Existing approaches \cite{yokoyama2024vlfm, zhou2023esc} often rely on greedy goal selection strategies, which choose the most promising viewpoint at each step but fail to account for long-term planning. This shortsighted decision-making can lead to inefficient search behaviors in complex environments.

% \begin{figure}
%     \centering
%     \includegraphics[width=1\linewidth]{./fig/motivation.png}
%     \caption{Motivation of our work Zero-Shot Object Navigation in dealing with object-based navigation scenario without prior knowledge of the environment.}
%     \label{fig:Motivation}
% \end{figure}
To overcome these challenges, we propose WTRP-Searcher, a novel training-free navigation framework that formulates ZSON as a Weighted Traveling Repairman Problem (WTRP) to optimize search efficiency through global viewpoint selection, as illustrated in 
\autoref{fig:teaser}. Our approach leverages a VLM to compute similarity scores between the environment and object descriptions, projecting them onto a 2D depth-enhanced map for improved spatial reasoning. Frontier-based viewpoint extraction assigns these similarity scores as weights in the WTRP optimization process, enabling more informed goal selection. Additionally, an open-vocabulary object detector dynamically refines candidate viewpoints and navigation targets based on detection confidence, while a 3D embedding feature map continuously updates environment memory, enhancing object localization as the agent explores.
%To address these problems, we propose WTRP-Searcher, which formulates object goal navigation as a Weighted Traveling Repairman Problem (WTRP) minimizing the weighted summary of viewpoints' waiting time. We utilize VLM to get the similarity between surroundings and object descriptions and project it to a 2D map with depth images to score viewpoints extracted from frontiers. These scores work as weights in WTRP. Current motion state and path length between agent and viewpoints relate to waiting time in WTRP. We deploy an open-vocabulary detector to detect targets. Detected objects would be set as candidate viewpoints or navigation goals according to their confidence. Along the exploration and searching procedure, we would build a 3D embedding feature map to match environments and targets simplifying the object search task later. 

In general, the contributions of this work are as follows:

\begin{itemize}
\item We propose WTRP-Searcher, a training-free framework for Zero-Shot Object Navigation (ZSON) that integrates open-vocabulary detection, 3D feature mapping, and a Weighted Traveling Repairman Problem (WTRP) formulation for efficient global planning for object search.

\item We introduce a global goal selection policy leveraging items and global associations, addressing the limitations of greedy strategies in prior work.
\item We develop a real-time mapping system that builds multi-modal maps, enabling efficient object localization and environment memorization.
\item We evaluate our method in simulated and real-world environments, showing superior performance over state-of-the-art approaches. We will open source our work for the benefit of society.
%, especially in multi-object tasks, with improvements of 5.6\% in SR and 15.2\% in SPL.
\end{itemize}

% \begin{figure*}[t]
%     \centering
%     \includegraphics[width=0.90\linewidth]{fig/real.png}
%     \caption{Implementation of our approach in a real robot. The left image visualizes items that support the search framework. The right top image represents the detection result of a camera in the robot. The right bottom image is a third view.}
%     \label{fig:real}
% \end{figure*}

\section{RELATED WORKS}

\subsection{Frontier-based Navigation}
Frontier-based navigation is a widely used approach in exploration tasks, where frontiers mark the boundaries between explored and unexplored areas \cite{yamauchi1997frontier}. Classical methods often employ greedy strategies, selecting the nearest frontier at each step to maximize immediate exploration gains. More advanced strategies, such as next-best-view selection \cite{bircher2018receding}, evaluate candidate viewpoints based on information gain and path cost, selecting the one with the highest utility. To improve planning efficiency, recent works \cite{zhou2021fuel,zhao2023autonomous} have reformulated frontier selection as a Traveling Salesman Problem (TSP), aiming to minimize the total travel cost across all viewpoints, which has demonstrated superior performance compared to traditional heuristic approaches.

Beyond general exploration, frontier-based strategies have been widely adopted in object search problems. Extensions of classical frameworks \cite{papatheodorou2023finding,luo2024star} integrate unknown space exploration with object-centric surface inspection, enabling more target-aware navigation. Several works focus on leveraging frontier selection for object search optimization. For example, \cite{yu2023frontier} introduces a learning-based policy that generates navigation goals based on semantic maps and frontiers, while \cite{gadre2023cows} employs a simpler heuristic, iteratively selecting the closest frontier until an open-vocabulary detector confirms object detection. More recent approaches, such as VLF-M \cite{yokoyama2024vlfm}, combine visual-language similarity scoring with frontier selection, prioritizing viewpoints with the highest confidence to improve search efficiency.

This growing body of research highlights the effectiveness of frontier-based strategies in object search, yet many existing methods still suffer from greedy selection biases, lack of global planning, and reliance on 2D representations, which limits performance in complex, occluded, or long-horizon navigation tasks. Addressing these challenges remains a key direction for improving efficient and scalable object navigation.
% Frontier marks boundaries between explored and unexplored areas \cite{yamauchi1997frontier}. Approaches based on that are the most popular methods in classical exploration tasks. Some methods employ a greedy approach that selects the nearest frontier at each step. A next-best-view selection strategy \cite{bircher2018receding} evaluates viewpoints extracted from frontiers by information gain and path to the current pose and chooses the viewpoint with the highest score to visit.  Recent methods \cite{zhou2021fuel,zhao2023autonomous}  formulate the problem of visiting all viewpoints as a traveling salesman problem which minimizes the total length of the path to all viewpoints and have get better performance than methods before.

% These methods have been adopted in object search problems.\cite{papatheodorou2023finding,luo2024star} extend classical exploration framework to consider exploring unknown space and inspecting surface regions within distance threshold together. Many works focus on leveraging frontiers to improve specific object searches. \cite{yu2023frontier} trains a learning-based policy to generate navigation goals from frontiers and a semantic map. \cite{gadre2023cows} simply chooses the closest frontier as a navigation goal until the target object is detected by an open-vocabulary detector. \cite{yokoyama2024vlfm} build a value map from visual language similarity and choose the frontier with the highest confidence to visit. 

\subsection{Zero-Shot Object Navigation}
ZSON requires an agent to navigate toward a specified target without prior training or exposure to the object or environment. Traditional approaches can be broadly categorized into end-to-end learning methods and modular methods. End-to-end methods \cite{mirowski2016learning, ramrakhya2022habitat} employ deep reinforcement learning (RL) or supervised learning to directly predict navigation actions from raw sensory inputs. Meanwhile, modular methods \cite{chang2020semantic, chaplot2020object} construct multi-modal maps that facilitate high-level decision-making and low-level motion planning, improving interpretability and structured reasoning. However, both paradigms struggle with generalization, especially when encountering novel objects and unseen environments.

To address these limitations, recent works have explored foundation models for object navigation, leveraging commonsense knowledge from large-scale internet datasets to improve adaptability. Some approaches \cite{yu2023l3mvn, zhou2023esc, dorbala2023can} incorporate LLMs to generate text-based object queries or reason about object locations \cite{yuan2021survey}, enabling training-free exploration. Other methods \cite{majumdar2022zson, wei2024ovexp, zhao2023zero} integrate VLMs as additional input to learning-based frameworks, enhancing scene understanding and target recognition. Additionally, recent works \cite{yokoyama2024vlfm, lukas2024one} fuse VLMs with modular exploration methods, using semantic similarity scores to rank and prioritize navigation goals.

While these advances significantly improve zero-shot navigation capabilities, several challenges remain, including long-term spatial reasoning, efficient goal selection, and real-world deployment feasibility. Addressing these limitations is crucial for enabling robust, generalizable, and scalable ZSON solutions.
%Zero-shot object navigation involves an agent navigating to a specific target without any prior training or exposure to that object or environment. Classic methods for object navigation involve end-to-end methods \cite{mirowski2016learning,ramrakhya2022habitat } which directly generate navigation actions from sensor input by deep learning and reinforcement learning, and modular method \cite{chang2020semantic,chaplot2020object} which always build multi-modal map to support high-level and low-level planner for navigation. However, these methods all challenge in generalization.  Recent advances introduce foundation models into object navigation frameworks utilizing commonsense knowledge learned from internet-scale datasets to enhance adaptability to novel objects or environments. Some methods\cite{yu2023l3mvn,zhou2023esc,dorbala2023can} leverage a Large Language Model (LLM) that presents object detection in text format or enables commonsense reasoning for training-free exploration approaches. Some methods\cite{majumdar2022zson, wei2024ovexp,zhao2023zero} use features from a visual language model (VLM) as additional input to learning-based frameworks achieving improvement. In some works \cite{yokoyama2024vlfm, lukas2024one}, VLM also integrates with classic modular exploration methods scoring candidate points to guide navigation. 

\section{METHODOLOGY}
In this section, we present the system architecture and the proposed methods in detail, outlining how our approach integrates global planning, semantic understanding, and multi-modal mapping to achieve efficient zero-shot object navigation. 

\begin{figure*}[t]
    \centering
    \includegraphics[width=1.0\linewidth]{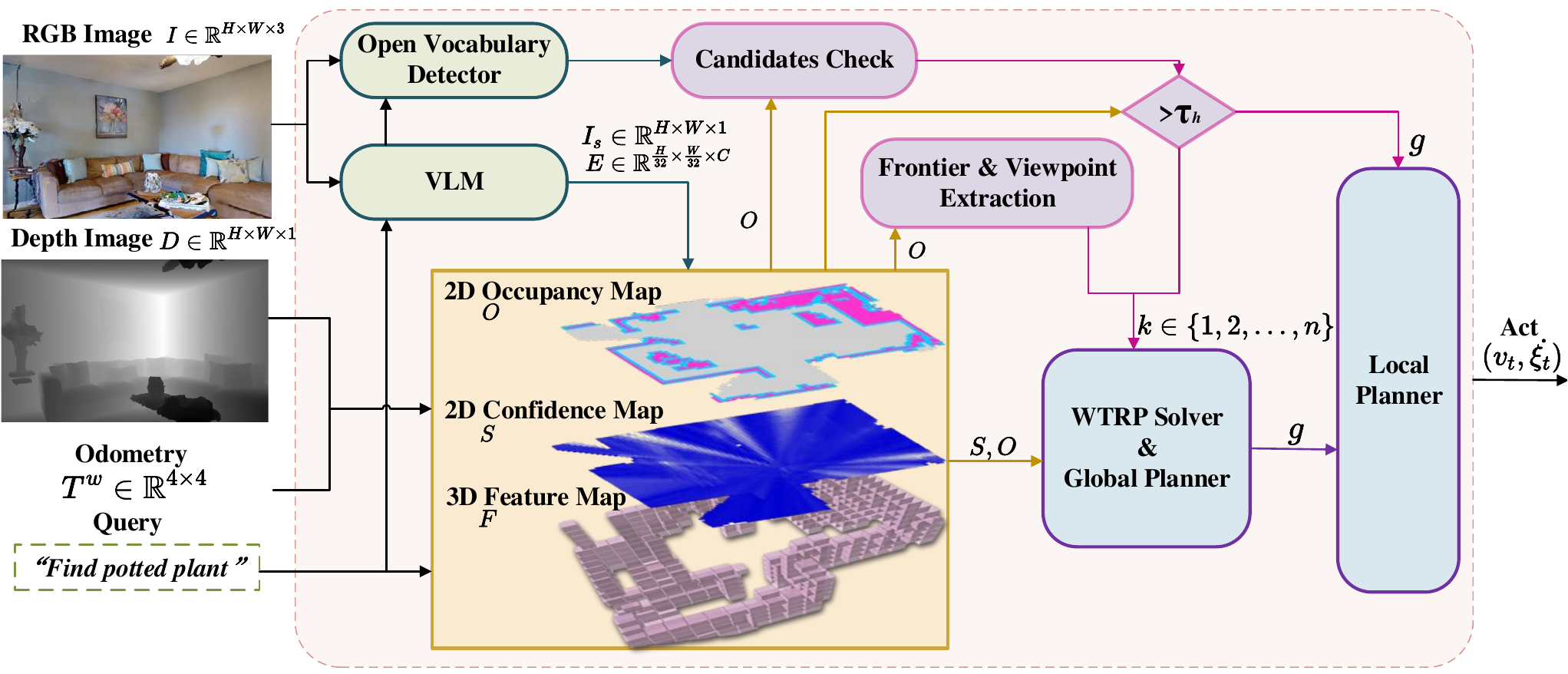}
    \caption{Overview of our framework. Sub-modules are colored and classified according to their functionality. VLM and open-vocabulary detector percept the environment and output features for mapping and planning; The mapping module constructs maps of three types from features, depth images, and poses; Modules in purple leverage detection results and multi-modal maps to generate candidate points for mid-term goals and potential navigation goals; All the candidate points are in overall planning by WTRP solver to get a global optimum route and the path to the first node would be sent to local planner to generate actions  }
    \label{fig:overview}
    \vspace{-10pt}
\end{figure*}

\subsection{Problem Formulation and System Overview}

%In a zero-shot object navigation task, the agent should navigate the environment to find objects that fulfill the requirement from flexible descriptions in a previously unknown environment. Besides the query, the input to the agent is a stream of rgb image, aligned depth image, and pose. The agent would generate a trajectory or a sequence of actions to target objects. \autoref{fig:overview} represents Our framework's architecture.
In a ZSON task, the agent must explore an unfamiliar environment to locate objects that match open-ended natural language descriptions. Unlike conventional navigation approaches that rely on predefined object categories or prior scene knowledge, zero-shot navigation requires the agent to interpret semantic queries dynamically and infer target locations in real time.

To achieve this, the agent receives an RGB image stream, an aligned depth image, and pose information as inputs. These sensory data are used to construct a spatial representation of the environment, allowing the agent to reason about potential object locations. Based on this representation, the agent plans a trajectory or generates a sequence of actions to navigate efficiently toward the target object. The overall architecture of our framework is illustrated in \autoref{fig:overview}, detailing the integration of multi-modal perception, mapping, and goal selection to enable efficient object search.

\subsection{Map Construction}

%Given a RGB image $I \in \mathbb{R}^{H \times W \times 3}$ we would extract features and project points to world space with an aligned depth image $I \in \mathbb{R}^{H \times W \times 1}$, camera information and pose $T \in \mathbb{R}^{4 \times 4}$ to build three types of maps: a 2D occupancy grid map for navigation and exploration, a 2D semantic confidence map to score candidate goals for planning, and a 3D feature map to match the environment with queries. 
Given an RGB image $I \in \mathbb{R}^{H \times W \times 3}$, we extract features and project points into world space using an aligned depth image $D \in \mathbb{R}^{H \times W \times 1}$, camera parameters, and the pose transformation matrix $T^w \in \mathbb{R}^{4 \times 4}$. This process constructs three complementary maps: a 2D occupancy grid map $O$ for navigation and exploration, a 2D semantic confidence map $S$ for goal selection, and a 3D feature map $F$ for object matching.
\subsubsection{Occupancy Grid Map}

%Each cell in the 2D occupancy grid map holds a probability value and inflation count indicating whether it is occupied, free, unknown, or inflated by occupied grids. The map is built by a classic method that we project depth points to 2D grids, use raycasting to determine whether a grid is hit or miss in this integration, and use the Bayesian method to update their state. After each integration, grids with a changed state would be used to update grids around it to inflate obstacles. 
Each cell in the 2D occupancy grid map holds a probability value and an inflation count, indicating whether it is occupied, free, unknown, or influenced by nearby occupied grids. The map is generated by projecting depth points onto a 2D grid, applying raycasting to classify grid cells as hits or misses, and updating their state using a Bayesian approach. After each update, surrounding grids are adjusted to reflect obstacle inflation, improving navigation safety.
\subsubsection{Semantic Confidence Map}
\label{sec:sim}

%Given a RGB image $I \in \mathbb{R}^{H \times W \times 3}$, we use connect \cite{liu2022convnet} based clip \cite{radford2021learning} to extract patch-level embeddings $E \in \mathbb{R}^{H/32 \times W/32 \times C}$ where $C$ is the dimension of clip embeddings. We compute the cosine similarity between embeddings of target descriptions and the image and then up-sample to image-level getting similarity image $I_s \in \mathbb{R}^{H \times W \times 1}$. Then we project these 2D points with aligned depth image $D \in \mathbb{R}^{H \times W \times 1}$ and pose $T \in \mathbb{R}^{4 \times 4}$ to 3D space. For points with the same index in x, y coordinates, we compute the average of the top 20\% similarity as assigned value for the grid in this observation denoting as $S_p^t$ to reduce the impact of irrelevant objects in stacking. The covariance of the value in the grid $p$ is:  
To integrate semantic awareness into navigation, we construct a 2D semantic confidence map using vision-language embeddings. Inspired by \cite{lukas2024one}, given an RGB image $I \in \mathbb{R}^{H \times W \times 3}$, we use a CLIP-based feature extractor \cite{liu2022convnet, radford2021learning} to obtain patch-level embeddings $E \in \mathbb{R}^{H/32 \times W/32 \times C}$, where $C$ denotes the embedding dimension. The cosine similarity between target descriptions and image patches is computed and then upsampled to the resolution of the input image, producing a per-pixel similarity representation, $I_s \in \mathbb{R}^{H \times W \times 1}$.

\begin{figure}[t]
    \centering
    \includegraphics[width=0.9\linewidth]{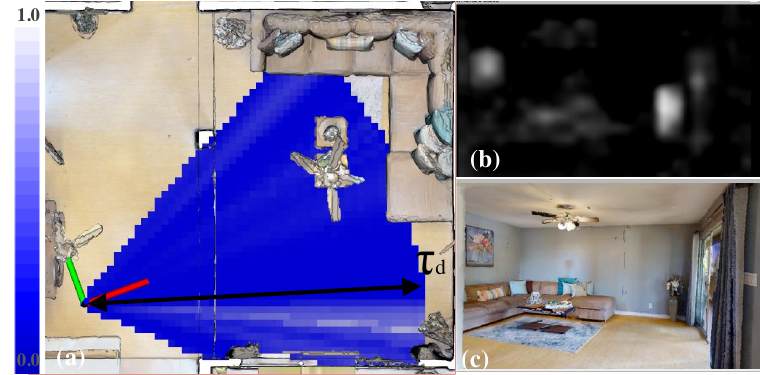}
    \caption{Construct a temporal semantic confidence map. (a) is the construction result. (b) is a pixel-wise confidence image generated from (c) by VLM. The mapping is limited in $\tau_d$ while points outside the region would also slightly contribute to the updating to enhance semantic guidance for navigation. }
    \label{fig:ray}
    \vspace{-10pt}
\end{figure}

These 2D similarity values are then projected into 3D world space using the aligned depth image $D \in \mathbb{R}^{H \times W \times 1}$ and the pose transformation matrix $T^w \in \mathbb{R}^{4 \times 4}$. To enhance robustness, the top 20\% similarity values are averaged at each spatial location, denoted as $S_p^t$. The covariance of the similarity value at grid cell $p$ is then computed as:

\begin{equation}
    \label{eq:cv}
   {\sigma_p^2}^t = \frac {\alpha_c * (d_{p,p_0})^2 }{ n_p},
\end{equation}
where $n_p$ is the number of points in the 2D grid, $d_{p,p_0}$ represents the distance of the grid to the sensor origin $p_0$, and $\alpha_c$ is a scaling factor. 

To maintain consistency over time, we employ a Kalman filter to update the similarity value $S_p$ in grid $p$ using the current observation $S_p^t$:
% where $n_p$ is the number of points in the 2D grid, $d_{p,p_0}$ is the distance of the grid to sensor origin $p_0$, $\alpha_c$ is a factor. 
% We use the Kalman filter to update the similarity value $S_p$ in grid $p$ with the current observation $S_p^t$:
\vspace{-5pt}
\begin{align}
    \label{eq:kal}
    K &= \frac{{\sigma_p^2}^t}{{\sigma_p^2}^- + {\sigma_p^2}^t},
   \\   
    S_p &= S_p^t + K (S_p^- - S_p^t),
\\
    {\sigma_p^2} &= (1 - K){\sigma_p^2}^t.
\end{align}

To evaluate viewpoints in free space, similarity values for these grids $p'$ are estimated by raycasting from the sensor to the grids hit by depth points $p$ :

% To evaluate viewpoints chosen in free space, it is important to score grids $p^{'} $ that are not hit by depth points. This is implemented by raycasting from hit grids $p$ to the sensor. The similarity of these grids is computed as:
\vspace{-5pt}
\begin{align}
    S_{p'}^t &= S_p^t - \gamma d_{p,p'}, \\
    \sigma_{p'}^2{}^t &= \beta \sigma_p^2{}^t,
\end{align}
where $\gamma$ and $\beta$ control similarity decay over distance. 

Additionally, to mitigate depth sensor noise, the updates are restricted to regions within a distance threshold $\tau_d$ from the sensor. \autoref{fig:ray} illustrates the process of constructing a temporal confidence map $S^t$ from an observation.

The confidence map is defined in a task-specific manner. For a newly specified target, the previous confidence map is reset. When multiple targets are searched simultaneously, an individual confidence map is updated and maintained for each target. The final confidence at each location is then obtained as the maximum across all confidence maps.

\subsubsection{3D Feature Map}

While the 2D confidence map is typically constructed for a specific task, maintaining a unified semantic memory of the environment would allow new tasks to leverage previously acquired knowledge, reducing search time and improving efficiency. Following the procedure described in \autoref{sec:sim}, at each time step $t$, we project feature embeddings into 3D space to form $F^t$ using aligned depth images and pose information. The average embedding $F_p^t \in F^t$ is computed for all embeddings mapped to the same 3D grid cell $p$. To maintain temporal consistency, we apply a Kalman filter to fuse $F_p^t$ with the previous feature map $F_p$, with the covariance and update equations following \autoref{eq:cv} and \autoref{eq:kal}, respectively. For efficient storage and traversal, we employ a modified octree structure \cite{duberg2020ufomap}, which exclusively stores occupied grids with embeddings. Additionally, grids corresponding to the floor and ceiling are systematically eliminated based on height thresholds, further optimizing the representation.

Since the VLM we employ produces patch-level features $E \in \mathbb{R}^{H/32 \times W/32 \times C}$, bilinear interpolation is typically used to obtain image-level features $E' \in \mathbb{R}^{H \times W \times C}$. Directly upsampling patch level features $E$ to pixel level features $E^{'}$, projecting it with aligned depth points from $D \in \mathbb{R}^{H \times W \times 1}$ to 3D grids, and then fusing features in the same voxel can be a straightforward method. 

% \begin{equation}
%  \mathcal{R}(x,y) =
% \left\{(u,v)\in\mathbb{Z}^2: 
% u \in \left\{\left\lfloor\frac{x}{r}\right\rfloor,\; \left\lfloor\frac{x}{r}\right\rfloor+1\right\},\;
% v \in \left\{\left\lfloor\frac{y}{r}\right\rfloor,\; \left\lfloor\frac{y}{r}\right\rfloor+1\right\}
% \right\}.   
% \end{equation}

\begin{equation}
\mathcal{R}(x, y) =
\left\{
(u, v) \in \mathbb{Z}^2 :
\begin{aligned}
&u \in \left\{\left\lfloor\frac{x}{r}\right\rfloor, \left\lfloor\frac{x}{r}\right\rfloor+1\right\}, \\
&v \in \left\{\left\lfloor\frac{y}{r}\right\rfloor, \left\lfloor\frac{y}{r}\right\rfloor+1\right\}
\end{aligned}
\right\}.
\end{equation}

\begin{equation}
    E'(x,y) =\sum_{(u,v)\in\mathcal{R}(x,y)}E(u,v)\cdot w_{u,v}(x,y),
\end{equation}

where the bilinear weights are
\begin{equation}
    w_{u,v}(x,y) =
    \begin{cases}
    \displaystyle
    \left(1 - \left| \frac{x}{r} - u \right|\right)
    \left(1 - \left| \frac{y}{r} - v \right|\right),
    & (u,v)\in\mathcal{R}(x,y), \\[10pt]
    0, & \text{otherwise.}
    \end{cases}
\end{equation}

\noindent Here \(r\) is the upsample factor, which is 32 in the scenario. $u,v$ are the related indices in $E$ of pixel $x,y$ in $E'$.

However, when the dimension $C$ is large, performing interpolation and fusion introduces significant computational and storage overhead. On the contrary, we integrate the prior-interpolation step with the fusion step internally and use the summarization of bilinear weights to replace the feature fusion in bilinear interpolation: when projecting the aligned depth points to 3D voxels, in each voxel $p$, we would count its related pixels in $E$ expressed as $\mathcal{D}_p$ and the corresponding weight. 
\begin{equation}
      \mathcal{W}^p_{u,v} = \sum_{(x,y)\in \mathcal{D}_p} w_{u,v}(x,y)  
\end{equation}

After the projection, we compute the current feature projected to voxel $p$ with the weighted average method.
\begin{equation}
    F^t_p = Norm(\frac{1}{\sum_{(u,v) \in \mathcal{R}(\mathcal{D}_p)}\mathcal{W}^p_{u,v} }\sum_{(u,v) \in \mathcal{R}(\mathcal{D}_p)}\mathcal{W}^p_{u,v}E(u,v))
\end{equation}

Because points projected to the same voxel are near, we simply assume they share the same covariance. When fusing $F^t_p$ with the previous observation $F_p$ following \autoref{eq:cv} and \autoref{eq:kal}, the $d_{p,p_0}$ is computed as the center of the voxel $p$ to the sensor. 
This strategy significantly reduces the number of high-dimensional embeddings processed during fusion.

\subsection{Candidate Generation Policy}
\label{seq:cand}
Candidate points for mid-term goals in object search are selected from frontier cluster viewpoints, detections from an open-vocabulary object detector, and feature map matching results.

In an unknown environment, frontiers are extracted from free-space grid cells adjacent to unknown regions with the occupancy map. Viewpoints are then generated by clustering these frontiers in previously visited free space, ensuring an optimal vantage point for exploration.

During navigation, an open-vocabulary detector continuously searches for target categories. We utilize YOLO-World \cite{cheng2024yolo} for real-time object detection. If the confidence score of a detected object exceeds the threshold $\tau_h$, its corresponding viewpoint is immediately set as the next navigation goal $g$. For objects with confidence scores between $\tau_l$ and $\tau_h$, the viewpoint is added as a candidate point for consideration in the WTRP solver. Since high-confidence detections can sometimes be false positives due to specific viewing angles or distances, detected targets are re-evaluated during navigation. If the confidence score drops below $\tau_h$ for a sustained period, the target is removed from the candidate list.

As the agent navigates, the 3D feature map is built and continuously updated. This map enhances the system’s environment memorization capability, improving target localization. Given a language-based query, we compute the similarity between the description and stored embeddings in each 3D grid. Similar to object detections, locations corresponding to high-confidence grid cells are directly treated as navigation goals $g$, while those with slightly lower confidence are incorporated into the WTRP solver for optimized selection.

\subsection{Mid-term Goal Choosing Policy}
This section describes the method to choose the mid-term goal when candidates and related information are given. Recent advancements in exploration tasks have formulated the problem of visiting all viewpoints as a Traveling Salesman Problem (TSP), aiming to minimize the length of the global tour. However, directly applying this formulation to object search tasks may not be optimal, as the agent stops once it finds a target—meaning that, in most cases, not all viewpoints are visited. Additionally, from a behavioral perspective, viewpoints with higher semantic similarity to the target should be prioritized over purely minimizing the travel distance. A global tour optimization that neglects semantic relevance may result in suboptimal search efficiency. 

Following \cite{afrati1986complexity}, we define the Weighted Traveling Repairman Problem (WTRP) as follows:

Given a set of points $k \in \{0,1,...,n\}$ with associated weights $W_k$, a cost matrix $M_{k_r, k_s}$ representing the time cost between any two points $k_r$ and $k_s$, and a start point $0 \in \{0,1,...,n\}$, the goal is to find a permutation $\pi_0=0, \pi_1, ..., \pi_{n}$ that minimizes:

\vspace{-5pt}
\begin{equation}
    \min_{\pi} \sum_{i=1}^{n} w_{\pi_{i}} \sum_{j=1}^{i} M_{\pi_{j-1}, \pi_{j}},
\end{equation}
where $\pi_j$ denotes the $j$-th node to be visited, $\pi_0$ is the agent's current position, and $W_{\pi_j}$ is the weight of the node, computed as:

% where $\pi_j $ denotes the j-th node to be visit, $\pi_0$ is current position of agent and 
$W_{\pi_j}$ denotes the weight of nodes and can be gained from:
\begin{align}
    W_{\pi_j} &= \phi(S_{p_{\pi_j}}),
    \\
    \phi(x) &=  \alpha_s e^{\mu x+\nu},  %\alpha_s e(10x-1),
\end{align}
where $\phi(x)$ is a mapping function designed to amplify the differences in node weights. The parameters $\alpha_s$, $\mu$, and $\nu$ are scaling factors.

Here, $p_k$ represents the position of a candidate or the agent in the map, and $S_{p_{\pi_j}}$ is the similarity of the node. If the candidate is the viewpoint from clusters, this can be retrieved from the 2D semantic confidence map based on its location. If the node is from the detection result of an open-vocabulary detector or the match result from the 3D feature map, $S_{p_{\pi_j}}$ is the detection confidence or match similarity, respectively.

Inspired by \cite{zhao2022faep, zhou2021fuel}, we incorporate both the distance between nodes and the agent's motion state—where $v_0$ and $\xi_0$ denote the agent's current velocity and orientation, while $v_{max}$ and $\dot{\xi}_{max}$ represent its maximum linear and angular velocities.$\xi_k$ describes the orientation of each candidate node.  The cost matrix is then defined as:

\begin{align}
    t(k_r,k_s) &= max\{ \frac{d_{p_{k_r},p_{k_s}}}{v_{max}},\frac{|\xi_{k_r}-\xi_{k_s} |}{\dot{\xi_{max}}} \},
    \\
    c_c(k) &= cos^{-1} \frac{(p_k-p_0)\cdot v_0}{||p_k-p_0|||v_0||},
    \\
    c_s(k) &= \frac{h_k}{h_{max}},
    \\
    M_{0,k} &= 
            t(0,k) + w_c\cdot c_c(k)+w_f\cdot c_s(k), \notag \\ &k \in {1,2,..., n},
     \\
    M_{k_r,k_s} &= M_{k_s,k_r} = t(k_r,k_s), \notag \\ &k_r,k_s \in {1,2,..., n},
    \\
    M_{k,0} &= 0,
\end{align}
where $t(k_r,k_s)$ denotes the travel time between nodes $k_r$ and $k_s$, while $c_c(k)$ ensures motion consistency. The term $c_s(k)$ is introduced to encourage the agent to first explore small, enclosed unexplored areas, which often result from incomplete observations and could otherwise cause inefficient back-and-forth maneuvers. This item is only used for candidates of viewpoints from clusters. 

To compute $h_k$, a ray is cast from the viewpoint toward the cluster center of its frontiers. If the ray intersects observed grids within the distance threshold $h_{max}$, $h_k$ is set as the distance from the hit grid to the cluster center. Otherwise, $h_k$ defaults to $h_{max}$.

We solve the WTRP using the LKH solver \cite{tinos2018efficient}. The first node in the computed tour serves as the mid-term goal and would be sent to the local planner \autoref{sec:local}. If the agent reaches the goal with candidates generated in the tour, or if the execution time exceeds the replan threshold $\tau_t$, the global tour is recomputed to ensure up-to-date planning; otherwise, the agent traverses the nodes of the tour in sequence.

\subsection{Local Planner}
\label{sec:local}
% Once the navigation goal is selected, according to the distance to the current position we would send waypoints generated by the A-star algorithm or the single point to the local planner. We modify \cite{zhou2022swarm} as a local planner utilizing MINCO \cite{wang2022geometrically} for trajectory optimization. Finally, a low-level controller analyzes the trajectories and generates velocity commands for the chassis. The local planner can handle dynamic obstacles and generate smooth trajectories so that improves the system's robustness and performance. 
Once a navigation goal $g$ is selected, a suitable path is generated based on the distance from the agent’s current position. If the target is farther away, waypoints are computed using the A* algorithm, whereas single-point goals are directly sent to the local planner. For trajectory optimization, we adapt the local planner from \cite{zhou2022swarm}, leveraging MINCO \cite{wang2022geometrically} to ensure smooth motion planning. The final linear and angular velocity commands $(v_t, \dot{\xi_t})$ for the chassis are generated by a low-level controller. The local planner is capable of handling dynamic obstacles and producing smooth, collision-free trajectories, thereby enhancing the system’s robustness and overall performance.

\begin{table}[t]
  \caption{Comparison with baselines in single-object navigation.}
  \centering
  \renewcommand{\arraystretch}{1.2}
  \setlength{\tabcolsep}{6pt}
  \resizebox{\linewidth}{!}{%
  \begin{tabular}{lcccc}
        \hline
          \hline
    \toprule
    \multirow{2}{*}{Method} & \multirow{2}{*}{Training-free} & \multirow{2}{*}{Foundation Model} & \multicolumn{2}{c}{HM3D} \\
    \cmidrule(l{1pt}r{1pt}){4-5}
    & & & SR$\uparrow$(\%) & SPL$\uparrow$(\%)  \\
    \midrule
    ZSON\cite{majumdar2022zson}   &  {\color{red}$\times$} & CLIP & 25.5 & 12.6  \\
    Voronav\cite{wu2024voronav}  &  {\color{green}$\checkmark$} & GPT-3.5 & 42.0  & 26.0  \\
    ESC\cite{zhou2023esc}     &  {\color{green}$\checkmark$} & GLIP, GPT-3.5 & 39.2\ & 22.3  \\
    VLFM\cite{yokoyama2024vlfm}    &  {\color{red}$\times$} & BLIP2 & 52.5  & 30.4  \\
    OneMap\cite{lukas2024one} &  {\color{green}$\checkmark$}     & CLIP &  55.8   &    37.4   \\
    Ours   &  {\color{green}$\checkmark$} & CLIP & \textbf{59.2}  & \textbf{39.8}  \\
    \bottomrule
  \end{tabular}%
  }
  \vspace{-10pt}
  \label{tab:single}
\end{table}

% \begin{table}[t]
%   \caption{Comprison with Baselines in single object navigation.}

% \vskip 0.05in

%   \centering
%   \resizebox{0.6\textwidth}{!}{%
%   \begin{tabular}{lccccc}
    
%     \toprule
%     \multirow{2}{*}{Method}  &\multirow{2}{*}{Training-free} & \multirow{2}{*}{Foundation Model} & \multicolumn{2}{c}{HM3D}   \\
%      \cmidrule(l{1pt}r{1pt}){4-5}
%     &&& SR$\uparrow$ & SPL$\uparrow$  \\
%     \midrule
%     ZSON &  $\times $ & CLIP & 0.255 & 0.126 &   \\
%      Voronav &  $\checkmark$ & GPT-3.5 & 0.420  & 0.260 & \\
%      ESC  &  $\checkmark$ & GLIP, GPT-3.5 & 0.392  & 0.223 &  \\

%    VLFM &  $\times$ & BLIP2 & 0.525  & 0.304  \\%topological

%      Ours & $\checkmark$ &CLIP & \textbf{0.592}  & \textbf{0.389} \\

%     \bottomrule

%   \end{tabular}%
%    }
% \vspace{-10pt}
%   \label{tab:single}
% \end{table}

\section{EXPERIMENTS}

% In this section, we first evaluate our method using the habitat simulator on HM3D dataset \cite{ramakrishnan2021habitat} with baselines in single and multiple object navigation. Then we carry out real-world experiments.
In this section, we evaluate our method using the Habitat simulator on the HM3D dataset \cite{ramakrishnan2021habitat} by comparing it with baseline methods in both single-object and multi-object navigation tasks. We further validate our approach through real-world experiments.

\subsection{Single Object Navigation}

% This task requires the agent to find an object of the target category in unknown environments. We evaluate methods in HM3D's validation split containing 2000 single-object episodes in 20 scenes across 6 categories.
This task requires the agent to locate an object belonging to the target category in an unknown environment. We evaluate our method using the validation split of the HM3D dataset, which consists of 2000 single-object navigation episodes across 20 scenes and 6 object categories.

\label{sec:single}
% {\bf Evaluation Metrics:} For all approaches, we report Success Rate (SR) and Success weighted by Path Length (SPL) which is the optimal path divided by actual path length weighted by a binary indicator of success. 
{\bf Evaluation Metrics:} We use standard evaluation metrics, including Success Rate (SR)  in \% and Success weighted by Path Length (SPL)  in \%. SR measures the proportion of successful episodes, while SPL accounts for both success and path efficiency, computed as the ratio of the optimal path length to the actual path length, weighted by a binary success indicator.

{\bf Baseline Selection:} We compare our approach against the current state-of-the-art (SOTA) methods in zero-shot object navigation, including ZSON \cite{majumdar2022zson}, VoroNav \cite{wu2024voronav}, ESC \cite{zhou2023esc}, VLFM \cite{yokoyama2024vlfm} and Onemap \cite{lukas2024one}. ZSON is a learning-based method that incorporates VLM embeddings as input. VoroNav constructs a Voronoi graph from a semantic map, enabling a large language model (LLM) to reason about waypoints. ESC and VLFM both employ frontiers for navigation: ESC utilizes a VLM for scene understanding and an LLM for goal selection, while VLFM ranks frontiers based on VLM similarity scores, selecting the highest-scoring frontier as the next waypoint. OneMap uses patch-level CLIP and constructs a 2D feature map storing VLM embeddings.

% {\bf Results Discussion:} The results of our comparison are presented in \autoref{tab:single}. Our method achieves superior performance over all zero-shot SOTA baselines in both SR and SPL. Specifically, we obtain an SR of 0.592 and an SPL of 0.389, representing a 12.7\% improvement in SR and a 27.9\% improvement in SPL compared to VLFM. 

% We attribute our higher SR to our candidate selection policy, where rechecking detected objects reduces the false positive rate, leading to more reliable target identification. Our significant SPL improvement indicates that formulating the route planning problem as a WTRP leads to more efficient trajectories compared to frontier-based approaches that greedily select the highest-scoring frontier. Additionally, the use of patch-level CLIP embeddings enables a more detailed and structured understanding of the environment than image-level visual-language models, allowing the constructed semantic similarity and feature map to guide exploration more effectively.

\begin{figure}[t]
    \centering
    \includegraphics[width=1\linewidth]{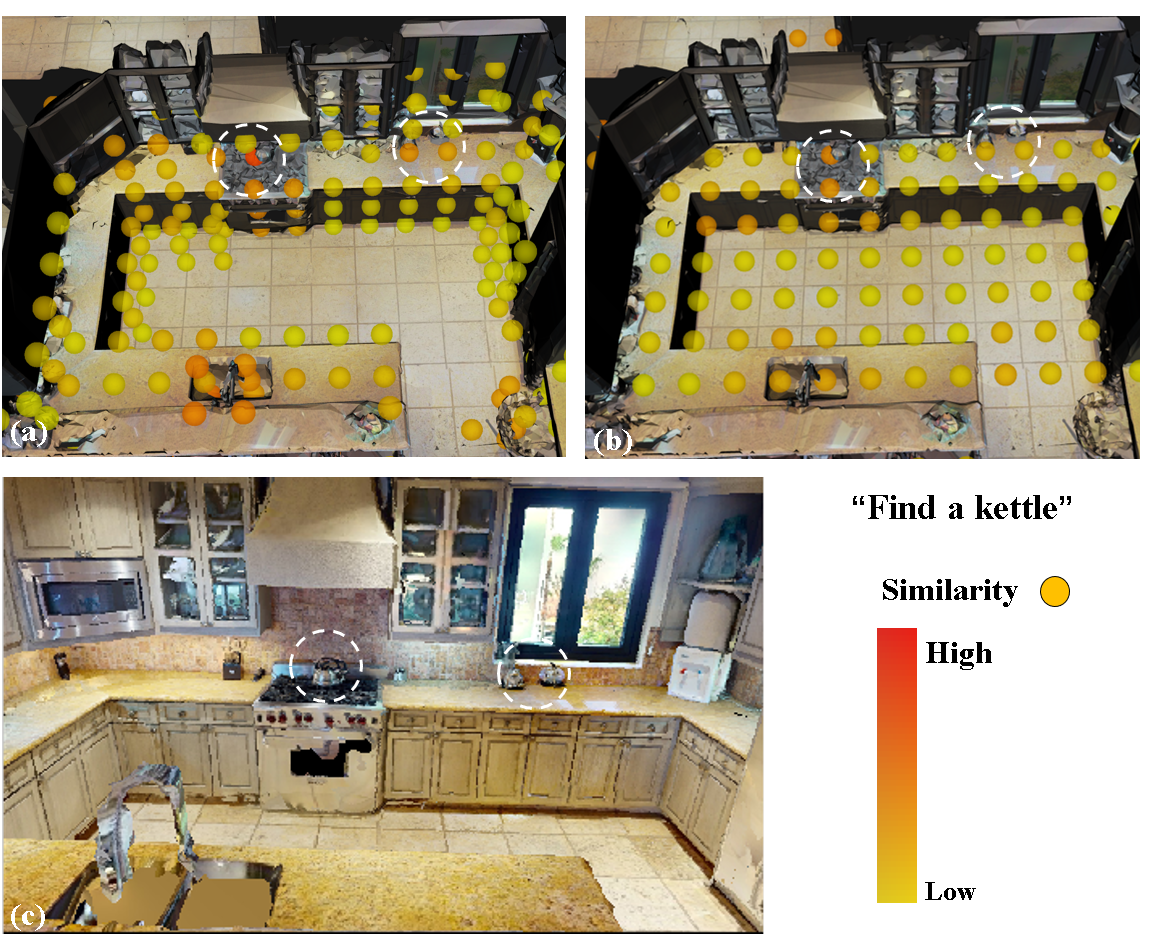}
    \caption{Comparison of 3D and 2D feature map. (a),(b)  is the result of 3D and 2D feature map, respectively; the colored spheres indicate the likelihood of target presence, where red denotes higher confidence regions.
    (c) is the scene, where targets are marked with a white circle in the figure.}
    \label{fig:3Dmap}
\end{figure}

\begin{table}[t]
  \caption{Multi-objet navigation evaluation}

\vskip 0.05in

  \centering
  \resizebox{0.48\textwidth}{!}{%
  \begin{tabular}{lccccc}
      \hline
          \hline
    \toprule
    \multirow{2}{*}{Method}& \multirow{2}{*}{Type} &\multicolumn{4}{c}{HM3D}  \\
         \cmidrule(l{1pt}r{1pt}){3-6}
  & & SR$\uparrow$(\%) & SPL$\uparrow$(\%)  & PR$\uparrow$(\%) &PPL$\uparrow$(\%)\\
    \midrule
    TSP Exp &\multirow{3}{*}{Simultaneous}& 29.3 & 12.1 & 49.3 & 17.8 \\
   % W/o CGP & & 35.1 & 15.4 & 57.5 & 19.4 \\
    W/o WTRP & & 52.1 &17.1& 67.8 &22.4  \\
    Ours  & &\textbf{53.4}& \textbf{23.0} & \textbf{68.9} & \textbf{28.9}\\
      \midrule
    OneMap\cite{lukas2024one} &\multirow{2}{*}{Sequential} &54.2&27.8&65.5&32.7 \\
    Ours & &\textbf{57.2}&\textbf{33.4}&\textbf{67.1}&\textbf{38.2}  \\

    \bottomrule
  \end{tabular}%
    }
    % \vspace{-20pt}
  \label{tab:multi}
\end{table}

\subsection{Multi-Object Navigation}

\begin{figure*}[t]
	\centering
     \includegraphics[width=0.99\linewidth]{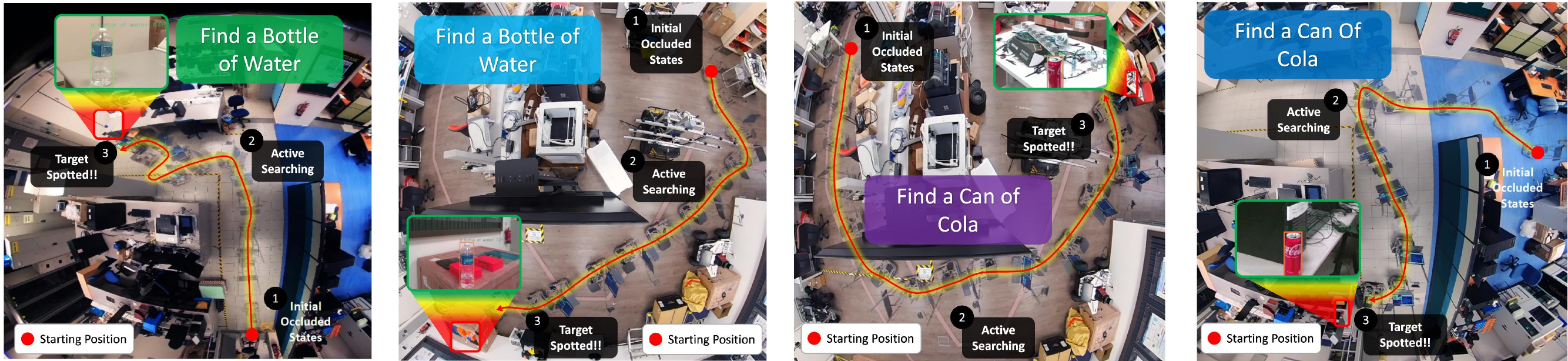}  %fig/real2(1).pdf is another version
	\caption{The process of visual language navigation in two real office scenes over two categories.}
	\label{fig:1234}
    \vspace{-15pt}
\end{figure*}

For multi-object navigation tasks, we design two types of experiments. In the first setting, the agent is given a list of target categories at the beginning and is required to find at least one object for each category. In the second setting, the agent receives the target categories sequentially, where a new target is assigned only after the current one is found. We evaluate our approach on the benchmark from \cite{lukas2024one}, which consists of 236 episodes in 20 scenes across six categories. Each episode includes three target categories, and the agent has no prior knowledge of the scene.

{\bf Evaluation Metrics:} In addition to the metrics defined in \autoref{sec:single}, we introduce additional metrics for multi-object navigation:  
Progress Rate (PR) in \%, which measures the rate at which the agent makes progress toward finding all target objects, and Progress Weighted by Path Length (PPL)  in \%, defined as:

\[
\text{PPL} = \frac{1}{N} \sum_{i=1}^{N} r_i \frac{l_i}{\max(p_i, l_i)},
\]
where $N$ is the number of episodes, $l_i$ is the optimal path length for episode $i$, and $r_i$ and $p_i$ represent the PR and the actual total path length of episode $i$, respectively.

{\bf Baseline Selection:} We compare our method with OneMap \cite{lukas2024one}, which is designed for sequential multi-object navigation. As no recent open-source approaches are available for simultaneous multi-object navigation, we conduct an ablation study to evaluate the impact of our framework's key components. Specifically, it includes replacing the WTRP solver with a greedy strategy ("W/o WTRP"), %eliminating candidates for mid-term goal choosing from detection and match results ("W/o CGP"), 
and adopting a TSP-based exploration framework that explores and detects the environment separately ("TSP Exp"), as shown in \autoref{tab:multi}.

\subsection{Results Discussion}

% {\bf Results Discussion:} The results in \autoref{tab:multi} demonstrate the effectiveness of our core technical components. The superior performance over the other methods in simultaneous planning underscores the advantage of our WTRP formulation in facilitating global, long-term planning. Compared to OneMap’s sequential planning, WTRP-Searcher exhibits improvements in SPL (0.323 vs. 0.278) and PPL (0.354 vs. 0.327), highlighting the synergy between global WTRP optimization and the 3D feature map. 

The results of our comparison are presented in \autoref{tab:single} and \autoref{tab:multi}. Our method achieves superior performance over all zero-shot SOTA baselines in both SR and SPL. 

Specifically, we achieve SRs of 59.2\% and 56.8\% in single- and multi-object navigation, corresponding to SPL improvements of 6.1\% and 5.8\% over OneMap. We attribute our higher SR to our candidate selection policy, where rechecking detected objects reduces the false positive rate, leading to more reliable target identification. Additionally, the use of patch-level CLIP embeddings (used by ours and OneMap) enables a more detailed and structured understanding of the environment than image-level visual-language models (used by VLFM), allowing the constructed semantic confidence and feature map to guide exploration more effectively and precisely.

We achieve a 6.3\% SPL improvement in single-object navigation, and improvements of 20\% in SPL and 16.8\% in PPL in multi-object navigation over OneMap. These improvements indicate that formulating the route planning problem as a WTRP leads to more efficient trajectories compared to frontier-based approaches that greedily select the highest-scoring frontier. The ablation study in simultaneous multi-object navigation evaluation also supports that.

%  The results in \autoref{tab:multi} demonstrate the effectiveness of our core technical components. The superior performance over the other methods in simultaneous planning underscores the advantage of our WTRP formulation in facilitating global, long-term planning. Compared to OneMap’s sequential planning, WTRP-Searcher exhibits improvements in SPL (0.323 vs. 0.278) and PPL (0.354 vs. 0.327), highlighting the synergy between global WTRP optimization and the 3D feature map.

% As shown in \autoref{tab:single} and \autoref{tab:multi}, our method consistently outperforms zero-shot SOTA baselines across both single-target and multi-target settings. In the single-target experiments (\autoref{tab:single}) we achieve SR = 0.592 and SPL = 0.398, corresponding to a 6.1\% SR and 6.3\% SPL improvement over OneMap; the higher SR is primarily due to our candidate-selection policy, where rechecking detected objects reduces false positives and yields more reliable target identification. Across both settings, improvements in trajectory efficiency (SPL, PPL) arise from two complementary factors: (1) formulating route planning as a WTRP produces more efficient global trajectories than frontier-based or sequential planning, and (2) patch-level CLIP embeddings deliver finer-grained semantic representations than image-level VLMs, enabling the 3D feature map to guide exploration more precisely. This synergy explains the gains in simultaneous planning, where WTRP-Searcher improves SPL (0.323 vs. 0.278) and PPL (0.354 vs. 0.327) compared to OneMap.

\subsection{Qualitative Analysis}
In environments with stacked spatial layouts, 3D feature maps exhibit superior performance in small object localization. For instance, as shown in \autoref{fig:3Dmap}, in a kitchen with cabinets above and a stove below, the 3D feature map more accurately highlights the spatial region containing the target object, such as a kettle. The 3D feature map’s capability to store and match environmental embeddings enables the agent to efficiently revisit high-similarity regions, reducing redundant exploration.

\begin{table}
  \caption{Time consumption of sub-modules on ORIN. In a Desktop PC, the performance can be doubled. We show the worst-case performance. An asterisk (*) indicates that the sub-module may also be triggered by external signals.}
    \renewcommand{\arraystretch}{1.0}
    \centering
    \begin{tabular}{ lcc}
      \hline
        \hline
        \toprule
        \textbf{Task} & \makecell{\textbf{Time consumption}\\\textbf{(s)}} &  \makecell{\textbf{Desired interval}\\\textbf{(s)} }  \\
        \hline
        VLM & $0.091$   &  $0.125$ \\ 
     Open-vocabulary detector & $0.079$  &  $0.125$ \\ 
       2D grid map & $0.108 $  &  $0.125$ \\  
       2D confidence map & $0.129 $  &  $0.125$ \\  
       3D feature map & $0.589 $  &  $0.250$ \\         
       WTRP solver  & $0.010$ & $ 8.000*$  \\  
       Trajectory optimization  & $0.005$ & $ 0.500*$  \\ 
       
        \bottomrule
    \end{tabular}
    \label{tab:time}
    \vspace{-15pt}
\end{table}

\subsection{Real World Experiments}
\label{sec:app}

% We deploy our method on a differential drive chassis using a realsense D455 depth camera to provide rgbd image and we deploy rtabmap \cite{labbe2019rtab} for odometry utilizing rgbd image and imu data from D455 camera. The whole system is deployed in jetson orin nx 16 GB. We list the time consumption of each sub-module for one iteration measured by second and the desired updating interval measured by second of each sub-module in \autoref{tab:time}. It shows that our system is efficient and can be easily deployed in real robot system. Implementation can be seen in the video. 
We deploy our method on a differential-drive robot equipped with a RealSense D455 depth camera, which provides RGB-D images. For odometry estimation, we utilize RTAB-Map \cite{labbe2019rtab}, leveraging both RGB-D images and IMU data from the D455 camera. The entire system is implemented on a Jetson Orin NX 16GB platform.

The time consumption of each sub-module per iteration, along with its desired updating interval, is summarized in \autoref{tab:time}. These results demonstrate the efficiency of our system, highlighting its feasibility for real-world robotic deployment. Although 3D feature map looks a bit heavy for the embedded device, the module serves to match language with the environment when receiving queries, and the latency in construction does not influence search. \autoref{fig:1234} illustrates top-down trajectories across four distinct scenarios, demonstrating the agent's ability to navigate around objects effectively. Further implementation details can be found in the accompanying video.

\section{CONCLUSION}

In this work, we present WTRP-Searcher, a framework for ZSON that casts the task as a Weighted Traveling Repairman Problem (WTRP). The method combines frontier-based exploration, open-vocabulary detection, and 3D feature matching to select candidate viewpoints, with WTRP guiding global planning and a local planner ensuring obstacle-aware execution. Experiments show that WTRP-Searcher surpasses state-of-the-art baselines in efficiency, accuracy, and adaptability across both simulation and real-world tests.
These results highlight the value of uniting spatial reasoning, semantic cues, and global optimization for ZSON. Future work will investigate lifelong and domain adaptation to strengthen robustness and sim-to-real generalization.

\ifCLASSOPTIONcaptionsoff
  \newpage
\fi

\bibliographystyle{IEEEtran}
\bibliography{IEEEexample}

\end{document}